\documentclass[10pt,twocolumn,letterpaper]{article}

\usepackage{iccv}
\usepackage{times}
\usepackage{epsfig}
\usepackage{graphicx}
\usepackage{amsmath}
\usepackage{amssymb}
\usepackage{subcaption}
\usepackage{multirow}
\usepackage{booktabs}

\usepackage{color}
\usepackage{xcolor}
\definecolor{citecolor}{HTML}{0071bc}
\usepackage[pagebackref=true,breaklinks=true,colorlinks,citecolor=citecolor,bookmarks=false]{hyperref}

\iccvfinalcopy 

\def\iccvPaperID{7065} 

\begin{document}

\title{Meta-Baseline: Exploring Simple Meta-Learning for Few-Shot Learning}

\author{Yinbo Chen\\
UC San Diego\\
\and
Zhuang Liu\\
UC Berkeley\\
\and
Huijuan Xu\\
Penn State University\\
\and
Trevor Darrell\\
UC Berkeley\\
\and
Xiaolong Wang\\
UC San Diego\\
}

\maketitle

\begin{abstract}
Meta-learning has been the most common framework for few-shot learning in recent years. It learns the model from collections of few-shot classification tasks, which is believed to have a key advantage of making the training objective consistent with the testing objective.
However, some recent works report that by training for whole-classification, i.e. classification on the whole label-set, it can get comparable or even better embedding than many meta-learning algorithms. The edge between these two lines of works has yet been underexplored, and the effectiveness of meta-learning in few-shot learning remains unclear.
In this paper, we explore a simple process: meta-learning over a whole-classification pre-trained model on its evaluation metric. We observe this simple method achieves competitive performance to state-of-the-art methods on standard benchmarks. Our further analysis shed some light on understanding the trade-offs between the meta-learning objective and the whole-classification objective in few-shot learning.
Our code is available at \url{https://github.com/yinboc/few-shot-meta-baseline}.
\end{abstract}

\section{Introduction}
While humans have shown incredible ability to learn from very few examples and generalize to many different new examples, the current deep learning approaches still rely on a large scale of training data. To mimic this human ability of generalization, few-shot learning~\cite{fei2006one,vinyals2016matching} is proposed for training networks to understand a new concept based on a few labeled examples. While directly learning a large number of parameters with few samples is very challenging and most likely leads to overfitting, a practical setting is applying transfer learning: train the network on common classes (also called base classes) with sufficient samples, then transfer the model to learn novel classes with a few examples. 

The meta-learning framework for few-shot learning follows the key idea of learning to learn. Specifically, it samples few-shot classification tasks from training samples belonging to the base classes and optimizes the model to perform well on these tasks. A task typically takes the form of $N$-way and $K$-shot, which contains $N$ classes with $K$ support samples and $Q$ query samples in each class. The goal is to classify these $N \times Q$ query samples into the $N$ classes based on the $N \times K$ support samples. Under this framework, the model is directly optimized on few-shot classification tasks. The consistency between the objectives of training and testing is considered as the key advantage of meta-learning. Motivated by this idea, many recent works~\cite{sung2018learning,gidaris2018dynamic,sun2019meta,wang2019tafe,finn2017model,rusu2018metalearning,lee2019meta,xu2021constellation} focus on improving the meta-learning structure, and few-shot learning itself has become a common testbed for evaluating meta-learning algorithms.

However, some recent works find that training for whole-classification, i.e. classification on the whole training label-set (base classes), provides the embedding that is comparable or even better than many recent meta-learning algorithms. The effectiveness of whole-classification models has been reported in both prior works~\cite{gidaris2018dynamic,chen2018a} and some concurrent works~\cite{wang2019simpleshot,tian2020rethink}. Meta-learning makes the form of training objective consistent with testing, but why it turns out to learn even worse embedding than simple whole-classification? While there are several possible reasons, e.g. optimization difficulty or overfitting, the answer has not been clearly studied yet. It remains even unclear that whether meta-learning is still effective compared to whole-classification in few-shot learning.

In this work, we aim at exploring the edge between whole-classification and meta-learning by decoupling the discrepancies. We start with Classifier-Baseline: a whole-classification method that is similarly proposed in concurrent works~\cite{wang2019simpleshot,tian2020rethink}. In Classifier-Baseline, we first train a classifier on base classes, then remove the last fully-connected (FC) layer which is class-dependent. During test time, it computes mean embedding of support samples for each novel class as their centroids, and classifies query samples to the nearest centroid with cosine distance. We observe this baseline method outperforms many recent meta-learning algorithms.

In order to understand whether meta-learning is still effective compared to whole-classification, a natural experiment is to see what happens if we perform further meta-learning over a converged Classifier-Baseline on its evaluation metric (i.e. cosine nearest-centroid). As a resulting method, it is similar to MatchingNet~\cite{vinyals2016matching} or ProtoNet~\cite{snell2017prototypical} with an additional classification pre-training stage. We observe that meta-learning can still improve Classifier-Baseline, and it achieves competitive performance to state-of-the-art methods on standard benchmarks. We call this simple method Meta-Baseline. We highlight that as a method, all the individual components of Meta-Baseline have been proposed in prior works, but to the best of our knowledge, it has been overlooked that none of the prior works studies them as a whole. We further decouple the discrepancies by evaluating on two types of generalization: \textit{base class generalization} denotes performance on few-shot classification tasks from unseen data in the base classes, which follows the common definition of generalization (i.e. evaluated in the training distribution); and \textit{novel class generalization} denotes performance on few-shot classification tasks from data in novel classes, which is the goal of the few-shot learning problem. We observe that: (i) During meta-learning, improving base class generalization can lead to worse novel class generalization; (ii) When training Meta-Baseline from scratch (i.e. without whole-classification training), it achieves higher base-class generalization but much lower novel class generalization.

Our observations suggest that there could be a trade-off between the objectives of meta-learning and whole-classification. It is likely that meta-learning learns the embedding that works better for $N$-way $K$-shot tasks, while whole-classification learns the embedding with stronger class transferability. We find that the main advantage of training for whole-classification before meta-learning is likely to be improving class transferability. Our further experiments provide a potential explanation of what makes Meta-Baseline a strong baseline: by inheriting one of the most effective evaluation metrics of the whole-classification model, it maximizes the reusing of the embedding with strong class transferability. From another perspective, our results also rethink the comparison between meta-learning and whole-classification from the perspective of datasets. When base classes are collected to cover the distribution of novel classes, novel-class generalization should converge to base-class generalization and the strength of meta-learning may overwhelm the strength of whole-classification.

In summary, our contributions are as following:
\begin{itemize}
    \item We present a simple Meta-Baseline that has been overlooked in prior work. It achieves competitive performance to state-of-the-art methods on standard benchmarks and is easy to follow.
    \item We observe a trade-off between the objectives of meta-learning and whole-classification, which potentially explains the success of Meta-Baseline and rethinks the effectiveness of both objectives in few-shot learning.
\end{itemize}

\section{Related Work}

Most recent approaches for few-shot learning follow the meta-learning framework. The various meta-learning architectures for few-shot learning can be roughly categorized into three groups. \emph{Memory-based methods}~\cite{Sachin2017,munkhdalai2017rapid,santoro2016meta,mishra2018a,munkhdalai2017meta} are based on the idea to train a meta-learner with memory to learn novel concepts (e.g. an LSTM-based meta-learner). \emph{Optimization-based methods}~\cite{grant2018recasting,rusu2018metalearning} follows the idea of differentializing an optimization process over support-set within the meta-learning framework: MAML~\cite{finn2017model} finds an initialization of the neural network that can be adapted to any novel task using a few optimization steps. MetaOptNet~\cite{lee2019meta} learns the feature representation that can generalize well for a linear support vector machine (SVM) classifier. Besides explicitly considering the dynamic learning process, \emph{metric-based methods}~\cite{vinyals2016matching} meta-learn a deep representation with a metric in feature space. For example, Prototypical Networks~\cite{snell2017prototypical} compute the average feature for each class in support-set and classify query samples by the nearest-centroid method. They use Euclidean distance since it is a Bregman divergence. Relation Networks~\cite{sung2018learning} further generalizes this framework by proposing a relation module as a learnable metric jointly trained with deep representations. TADAM~\cite{oreshkin2018tadam} proposes to use a task conditioned metric resulting in a task-dependent metric space.

\begin{table*}
    \begin{center}
        \begin{tabular}{lccc}
            \toprule
            \textbf{Method} & \textbf{Whole-classification training} & \textbf{Meta-learning} & \textbf{Others} \\
            Matching Networks~\cite{vinyals2016matching} & no / yes (large models) & attention + cosine & FCE \\
            Prototypical Networks~\cite{snell2017prototypical} & no & centroid + Euclidean & - \\
            Baseline++~\cite{chen2018a} & yes (cosine classifier) & - & fine-tuning \\ 
            Meta-Baseline (ours) & yes & centroid + cosine ($*\tau$) & - \\
            \bottomrule
        \end{tabular}
    \end{center}
    \caption{\textbf{Overview of method comparison.} We summarize the differences between Meta-Baseline and prior methods.}
    \label{tab:methods}
\end{table*}

While significant progress is made in the meta-learning framework, some recent works challenge the effectiveness of meta-learning with simple whole-classification, i.e. a classification model on the whole training label-set. Cosine classifier~\cite{gidaris2018dynamic} and Baseline++~\cite{chen2018a} perform whole-classification training by replacing the top linear layer with a cosine classifier, and they adapt the classifier to a few-shot classification task of novel classes by performing nearest centroid or fine-tuning a new layer respectively. They show these whole-classification models can achieve competitive performance compared to several popular meta-learning models. Another recent work~\cite{dhillon2019baseline} studies on a transductive setting. Along with these baseline methods, more advanced meta-learning methods~\cite{sun2019meta, lee2019meta, xu2021constellation} are proposed and they set up new state-of-the-art results. The effectiveness of whole-classification is then revisited in two of the concurrent works~\cite{wang2019simpleshot,tian2020rethink} with improved design choices.

By far, the effectiveness of meta-learning compared to whole-classification in few-shot learning is still unclear, since the edge between whole-classification models and meta-learning models remains underexplored. The goal of this work is to explore the insights behind the phenomenons. Our experiments show a potential trade-off between the meta-learning and whole-classification objectives, which provides a more clear understanding of the comparison between both objectives for few-shot learning.

As a method, similar ideas to Classifier-Baseline are concurrently reported in recent works~\cite{wang2019simpleshot,tian2020rethink}. Unlike some prior works~\cite{gidaris2018dynamic,chen2018a}, Classifier-Baseline does not replace the last layer with cosine classifier during training, it trains the whole-classification model with a linear layer on the top and applies cosine nearest-centroid metric during the test time for few-shot classification on novel classes. The Meta-Baseline is meta-learning over a converged Classifier-Baseline on its evaluation metric (cosine nearest-centroid). It is similar (with inconspicuous and important differences as shown in Table~\ref{tab:methods}) to those simple and classical metric-based meta-learning methods~\cite{vinyals2016matching,snell2017prototypical}. The main purpose of Meta-Baseline in this paper is to understand the comparison between whole-classification and meta-learning objectives, but we find it is also a simple meta-learning baseline that has been overlooked. While every individual component in Meta-Baseline is not novel, to the best of our knowledge, none of the prior works studies them as a whole.

\begin{figure*}
    \begin{center}
        \includegraphics[width=.85\linewidth]{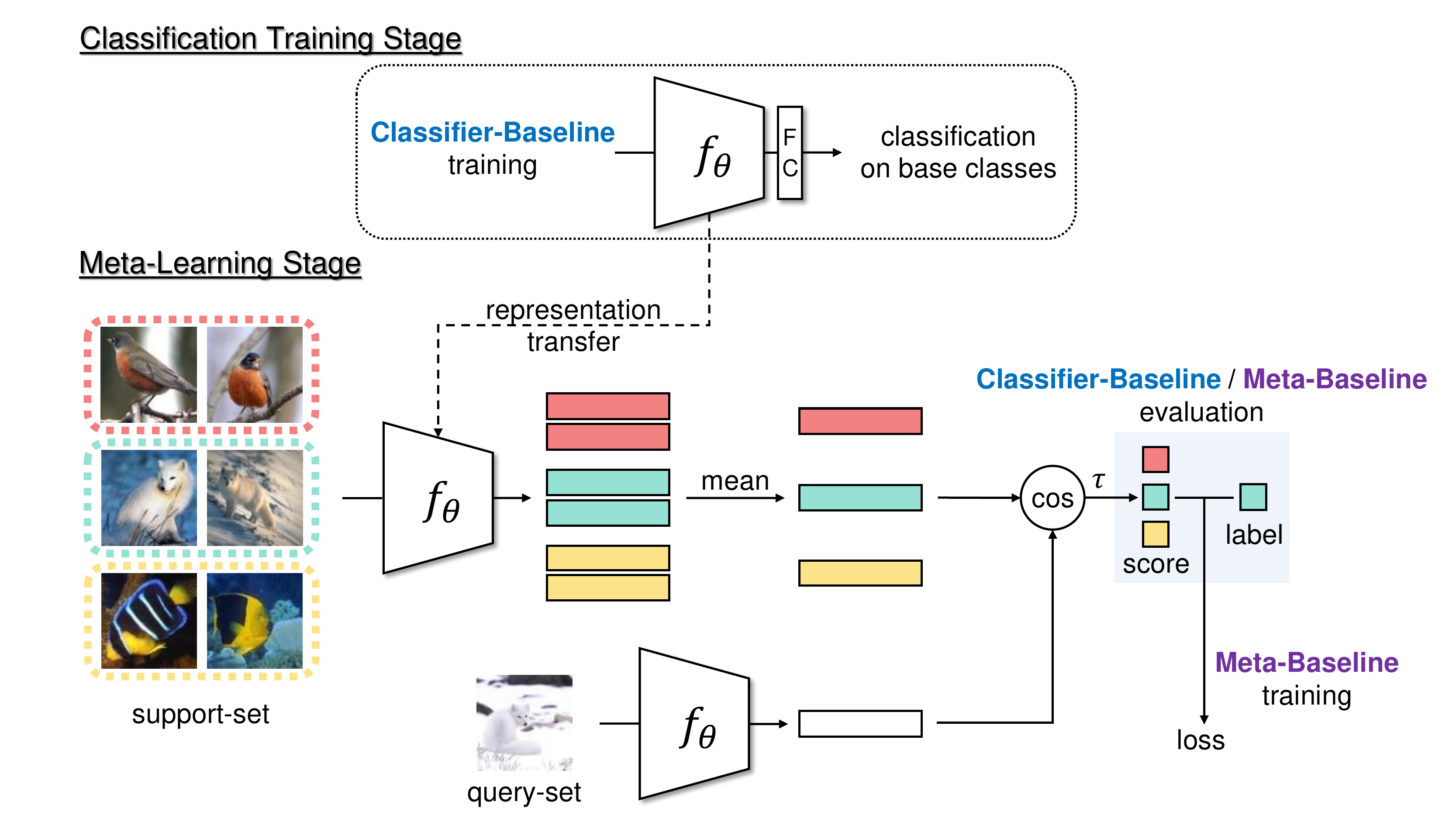}
    \end{center}
    \caption{\textbf{Classifier-Baseline and Meta-Baseline.} Classifier-Baseline is to train a classification model on all base classes and remove its last FC layer to get the encoder $f_{\theta}$. Given a few-shot task, it computes the average feature for samples of each class in support-set, then it classifies a sample in query-set by nearest-centroid with cosine similarity as distance. In Meta-Baseline, it further optimizes a converged Classifier-Baseline on its evaluation metric, and an additional learnable scalar $\tau$ is introduced to scale cosine similarity.}
    \label{fig:mb}
\end{figure*}

\section{Method}

\subsection{Problem definition}

In standard few-shot classification, given a labeled dataset of base classes $C_{base}$ with a large number of images, the goal is to learn concepts in novel classes $C_{novel}$ with a few samples. In an $N$-way $K$-shot few-shot classification task, the \textit{support-set} contains $N$ classes with $K$ samples per class, the \textit{query-set} contains samples from the same $N$ classes with $Q$ samples per class, and the goal is to classify the $N \times Q$ query images into $N$ classes.

\subsection{Classifier-Baseline}

Classifier-Baseline is a whole-classification model, i.e. a classification model trained for the whole label-set. It refers to training a classifier with classification loss on all base classes and performing few-shot tasks with the cosine nearest-centroid method. Specifically, we train a classifier on all base classes with standard cross-entropy loss, then remove its last FC layer and get the encoder $f_{\theta}$, which maps the input to embedding. Given a few-shot task with the support-set $S$, let $S_c$ denote the few-shot samples in class $c$, it computes the average embedding $w_c$ as the centroid of class $c$:
\begin{equation}
    w_c = \frac{1}{|S_c|} \sum_{x \in S_c} f_{\theta}(x),
\label{eq:baseline-wc}
\end{equation}
then for a query sample $x$ in a few-shot task, it predicts the probability that sample $x$ belongs to class $c$ according to the cosine similarity between the embedding of sample $x$ and the centroid of class $c$:
\begin{equation}
    p(y = c \mid x) = \frac{\exp \big(\langle f_{\theta}(x), w_c \rangle \big)}{\sum_{c'} \exp \big(\langle f_{\theta}(x), w_{c'} \rangle \big)},
\label{eq:baseline-pyx}
\end{equation}
where $\langle \cdot, \cdot \rangle$ denotes the cosine similarity of two vectors.

Similar methods to Classifier-Baseline have also been proposed in concurrent works~\cite{wang2019simpleshot,tian2020rethink}. Compared to Baseline++~\cite{chen2018a}, the Classifier-Baseline does not use the cosine classifier for training or perform fine-tuning during testing, while it performs better on standard benchmarks. In this work, we choose Classifier-Baseline as the representative of whole-classification models for few-shot learning. For simplicity and clarity, we do not introduce additional complex techniques for this whole-classification training.

\subsection{Meta-Baseline}

Figure \ref{fig:mb} visualizes the Meta-Baseline. The first stage is the classification training stage, it trains a Classifier-Baseline, i.e. training a classifier on all bases classes and remove its last FC layer to get $f_{\theta}$. The second stage is the meta-learning stage, which optimizes the model on the evaluation metric of Classifier-Baseline. Specifically, given the classification-trained feature encoder $f_{\theta}$, it samples $N$-way $K$-shot tasks (with $N \times Q$ query samples) from training samples in base classes. To compute the loss for each task, in support-set it computes the centroids of $N$ classes defined in Equation \ref{eq:baseline-wc}, which are then used to compute the predicted probability distribution for each sample in query-set defined in Equation \ref{eq:baseline-pyx}. The loss is a cross-entropy loss computed from $p$ and the labels of the samples in the query-set. During training, each training batch can contain several tasks and the average loss is computed.

Since cosine similarity has the value range of $[-1, 1]$, when it is used to compute the logits, it can be helpful to scale the value before applying Softmax function during training (a common practice in recent work~\cite{gidaris2018dynamic,qi2018low,oreshkin2018tadam}). We multiply the cosine similarity by a learnable scalar $\tau$, and the probability prediction in training becomes:
\begin{equation}
    p(y = c \mid x) = \frac{\exp \big(\tau \cdot \langle f_{\theta}(x), w_c \rangle \big)}{\sum_{c'} \exp \big(\tau \cdot \langle f_{\theta}(x), w_{c'} \rangle \big)}.
\label{eq:baseline-pyx-ex}
\end{equation}

In this work, the main purpose of Meta-Baseline is to investigate whether the meta-learning objective is still effective over a whole-classification model. As a method, while every component in Meta-Baseline has been proposed in prior works, we find none of the prior works studies them as a whole. Therefore, Meta-Baseline should also be an important baseline that has been overlooked.

\begin{table*}
    \begin{center}
        \begin{tabular}{llcc}
            \toprule
            \textbf{Model} & \textbf{Backbone} & \textbf{1-shot} & \textbf{5-shot} \\
            \midrule
            Matching Networks \cite{vinyals2016matching} & ConvNet-4 & 43.56 $\pm$ 0.84 & 55.31 $\pm$ 0.73 \\
            Prototypical Networks \cite{snell2017prototypical} & ConvNet-4 & 48.70 $\pm$ 1.84 & 63.11 $\pm$ 0.92 \\
            Prototypical Networks (re-implement) & ResNet-12 & 53.81 $\pm$ 0.23 & 75.68 $\pm$ 0.17 \\
            Activation to Parameter \cite{qiao2018few} & WRN-28-10 & 59.60 $\pm$ 0.41 & 73.74 $\pm$ 0.19 \\
            LEO \cite{rusu2018metalearning} & WRN-28-10 & 61.76 $\pm$ 0.08 & 77.59 $\pm$ 0.12 \\
            Baseline++ \cite{chen2018a} & ResNet-18 & 51.87 $\pm$ 0.77 & 75.68 $\pm$ 0.63 \\
            SNAIL \cite{mishra2018a} & ResNet-12 & 55.71 $\pm$ 0.99 & 68.88 $\pm$ 0.92 \\
            AdaResNet \cite{munkhdalai2017rapid} & ResNet-12 & 56.88 $\pm$ 0.62 & 71.94 $\pm$ 0.57 \\
            TADAM \cite{oreshkin2018tadam} & ResNet-12 & 58.50 $\pm$ 0.30 & 76.70 $\pm$ 0.30 \\
            MTL \cite{sun2019meta} & ResNet-12 & 61.20 $\pm$ 1.80 & 75.50 $\pm$ 0.80 \\
            MetaOptNet \cite{lee2019meta} & ResNet-12 & 62.64 $\pm$ 0.61 & 78.63 $\pm$ 0.46 \\
            SLA-AG \cite{lee2020self} & ResNet-12 & 62.93 $\pm$ 0.63 & 79.63 $\pm$ 0.47 \\
            ProtoNets + TRAML \cite{li2020boosting} & ResNet-12 & 60.31 $\pm$ 0.48 & 77.94 $\pm$ 0.57 \\
            ConstellationNet \cite{xu2021constellation} & ResNet-12 & \textbf{64.89 $\pm$ 0.23} & \textbf{79.95 $\pm$ 0.17} \\
            \midrule
            Classifier-Baseline (ours) & ResNet-12 & 58.91 $\pm$ 0.23 & 77.76 $\pm$ 0.17 \\
            Meta-Baseline (ours) & ResNet-12 & 63.17 $\pm$ 0.23 & 79.26 $\pm$ 0.17 \\
            \bottomrule
        \end{tabular}
    \end{center}
    \caption{\textbf{Comparison to prior works on miniImageNet.} Average 5-way accuracy (\%) with 95\% confidence interval.}
    \label{tab:mini}
\end{table*}

\begin{table*}
    \begin{center}
        \begin{tabular}{llcc}
        \toprule
        \textbf{Model} & \textbf{Backbone} & \textbf{1-shot} & \textbf{5-shot} \\
        \midrule
        MAML \cite{finn2017model} & ConvNet-4 & 51.67 $\pm$ 1.81 & 70.30 $\pm$ 1.75 \\
        Prototypical Networks* \cite{snell2017prototypical} & ConvNet-4 & 53.31 $\pm$ 0.89 & 72.69 $\pm$ 0.74 \\
        Relation Networks* \cite{sung2018learning} & ConvNet-4 & 54.48 $\pm$ 0.93 & 71.32 $\pm$ 0.78 \\
        LEO \cite{rusu2018metalearning} & WRN-28-10 & 66.33 $\pm$ 0.05 & 81.44 $\pm$ 0.09 \\
        MetaOptNet \cite{lee2019meta} & ResNet-12 & 65.99 $\pm$ 0.72 & 81.56 $\pm$ 0.53 \\
        \midrule
        Classifier-Baseline (ours) & ResNet-12 & 68.07 $\pm$ 0.26 & \textbf{83.74 $\pm$ 0.18} \\
        Meta-Baseline (ours) & ResNet-12 & \textbf{68.62 $\pm$ 0.27} & \textbf{83.74 $\pm$ 0.18} \\
        \bottomrule
        \end{tabular}
    \end{center}
    \caption{\textbf{Comparison to prior works on tieredImageNet.} Average 5-way accuracy (\%) with 95\% confidence interval.}
    \label{tab:tiered}
\end{table*}

\begin{table*}
\begin{center}
\begin{tabular}{llcc}
\toprule
\textbf{Model} & \textbf{Backbone} & \textbf{1-shot} & \textbf{5-shot} \\
\midrule
Classifier-Baseline (ours) & ResNet-18 & 83.51 $\pm$ 0.22 & \textbf{94.82 $\pm$ 0.10} \\
Meta-Baseline (ours) & ResNet-18 & \textbf{86.39 $\pm$ 0.22} & \textbf{94.82 $\pm$ 0.10} \\
\midrule
Classifier-Baseline (ours) & ResNet-50 & 86.07 $\pm$ 0.21 & \textbf{96.14 $\pm$ 0.08} \\
Meta-Baseline (ours) & ResNet-50 & \textbf{89.70 $\pm$ 0.19} & \textbf{96.14 $\pm$ 0.08} \\
\bottomrule
\end{tabular}
\end{center}
\caption{\textbf{Results on ImageNet-800.} Average 5-way accuracy (\%) is reported with 95\% confidence interval.}
\label{tab:im800}
\end{table*}

\section{Results on Standard Benchmarks}

\subsection{Datasets}

The \emph{miniImageNet} dataset~\cite{vinyals2016matching} is a common benchmark for few-shot learning. It contains 100 classes sampled from ILSVRC-2012 \cite{russakovsky2015imagenet}, which are then randomly split to 64, 16, 20 classes as training, validation, and testing set respectively. Each class contains 600 images of size $84 \times 84$.

The \emph{tieredImageNet} dataset~\cite{ren2018meta} is another common benchmark proposed more recently with much larger scale. It is a subset of ILSVRC-2012, containing 608 classes from 34 super-categories, which are then split into 20, 6, 8 super-categories, resulting in 351, 97, 160 classes as training, validation, testing set respectively. The image size is $84 \times 84$. This setting is more challenging since \textit{base classes and novel classes come from different super-categories.}

In addition to the datasets above, we evaluate our model on \emph{ImageNet-800}, which is derived from ILSVRC-2012 1K classes by randomly splitting 800 classes as base classes and 200 classes as novel classes. The base classes contain the images from the original training set, the novel classes contain the images from the original validation set. This larger dataset aims at making the training setting standard as the ImageNet 1K classification task~\cite{he2016deep}.

\subsection{Implementation details}

We use ResNet-12 that follows the most of recent works~\cite{oreshkin2018tadam, sun2019meta, lee2019meta, xu2021constellation} on miniImageNet and tieredImageNet, and we use ResNet-18, ResNet-50~\cite{he2016deep} on ImageNet-800. For the \textit{classification training stage}, we use the SGD optimizer with momentum 0.9, the learning rate starts from 0.1 and the decay factor is 0.1. On miniImageNet, we train 100 epochs with batch size 128 on 4 GPUs, the learning rate decays at epoch 90. On tieredImageNet, we train 120 epochs with batch size 512 on 4 GPUs, the learning rate decays at epoch 40 and 80. On ImageNet-800, we train 90 epochs with batch size 256 on 8 GPUs, the learning rate decays at epoch 30 and 60. The weight decay is 0.0005 for ResNet-12 and 0.0001 for ResNet-18 or ResNet-50. Standard data augmentation is applied, including random resized crop and horizontal flip. For \textit{meta-learning stage}, we use the SGD optimizer with momentum 0.9. The learning rate is fixed as 0.001. The batch size is 4, i.e. each training batch contains 4 few-shot tasks to compute the average loss. The cosine scaling parameter $\tau$ is initialized as 10.

We also apply \emph{consistent sampling} for evaluating the performance. For the novel class split in a dataset, the sampling of testing few-shot tasks follows a deterministic order. Consistent sampling allows us to get a better model comparison with the same number of sampled tasks. In the following sections, when the confidence interval is omitted in the table, it indicates that a fixed set of 800 testing tasks are sampled for estimating the performance.

\begin{figure*}
    \begin{center}
        \includegraphics[width=\linewidth]{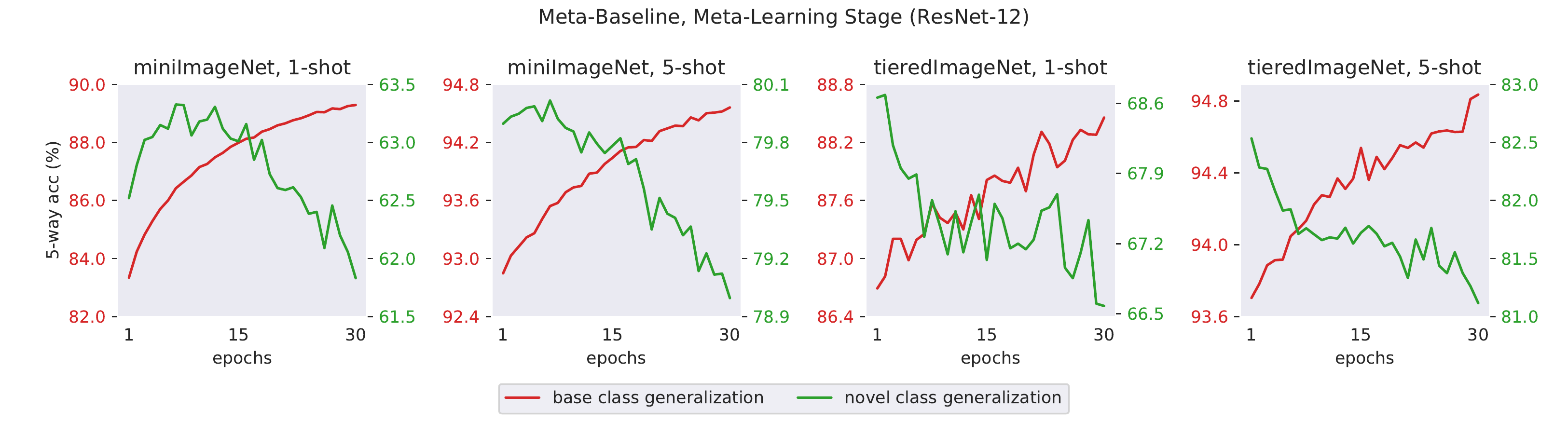}
    \end{center}
    \caption{\textbf{Objective discrepancy of meta-learning on miniImageNet and tieredImageNet.} Each epoch contains 200 training batches. Average 5-way accuracy (\%) is reported.}
    \label{fig:mbtr}
\end{figure*}

\begin{figure}
    \begin{center}
        \includegraphics[width=\linewidth]{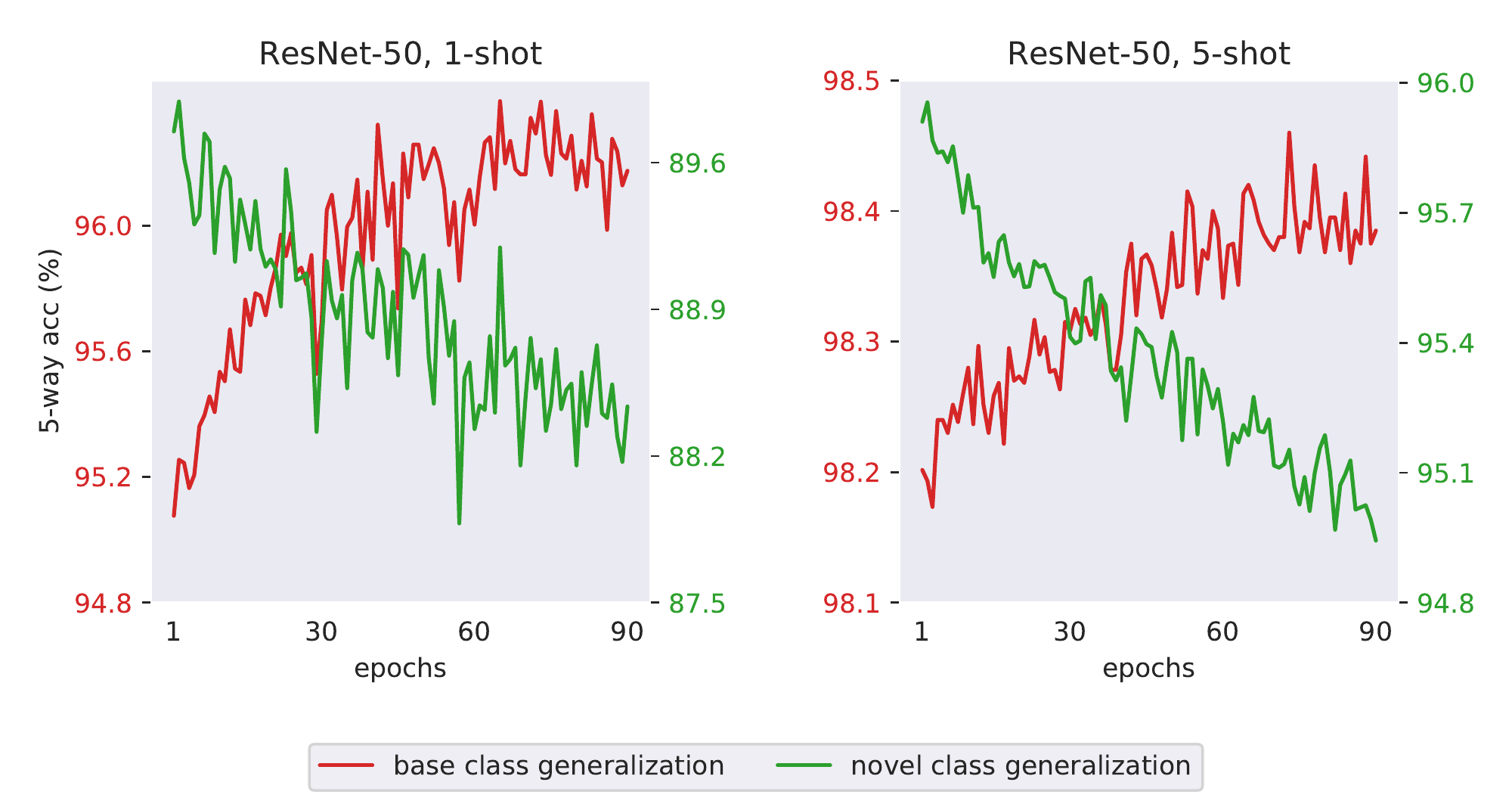}
    \end{center}
    \caption{\textbf{Objective discrepancy of meta-learning on ImageNet-800.} Each epoch contains 500 training batches. Average 5-way accuracy (\%) is reported.}
    \label{fig:mbtr-im800-mini}
    \vspace{-1em}
\end{figure}

\begin{table*}
    \begin{center}
        \begin{tabular}{llcccc}
            \toprule
            \textbf{Task} & \textbf{Model} & \textbf{mini-tiered} & \textbf{mini-shuffled} & \textbf{full-tiered} & \textbf{full-shuffled} \\
            \midrule
            \multirow{3}{*}{1-shot} & Classifier-Baseline & 56.91 & 61.64 & 68.76 & 77.67 \\
            & Meta-Baseline & 58.44 & 65.88 & 69.52 & 80.48 \\
            & $\Delta$ & \emph{+1.53} & \emph{+4.24} & \emph{+0.76} & \emph{+2.81} \\
            \midrule
            \multirow{3}{*}{5-shot} & Classifier-Baseline & 74.30 & 79.26 & 84.07 & 90.58 \\
            & Meta-Baseline & 74.63 & 80.58 & 84.07 & 90.67 \\
            & $\Delta$ & \emph{+0.33} & \emph{+1.32} & \emph{+0.00} & \emph{+0.09} \\
            \bottomrule
        \end{tabular}
    \end{center}
    \caption{\textbf{Effect of dataset properties.} Average 5-way accuracy (\%), with ResNet-12.}
    \label{tab:tvars}
\end{table*}

\begin{table*}
\begin{minipage}{0.475\linewidth}
    \centering
    \begin{tabular}{llcc}
    \toprule
    & \textbf{Training} & \textbf{Base gen.} & \textbf{Novel gen.} \\
    \midrule
    \multirow{2}{*}{1-shot} & w/ ClsTr & 86.42 & \textbf{63.33} \\
    & w/o ClsTr & \textbf{86.74} & 58.54 \\
    \midrule
    \multirow{2}{*}{5-shot} & w/ ClsTr & 93.54 & \textbf{80.02} \\
    & w/o ClsTr & \textbf{94.47} & 74.95 \\
    \bottomrule
    \end{tabular}
    \caption{\textbf{Comparison on Meta-Baseline training from scratch.} Average 5-way accuracy (\%), with ResNet-12 on miniImageNet. ClsTr: classification training stage.}
    \label{tab:mbtrs}
\end{minipage}
\hspace{0.04\linewidth}
\begin{minipage}{0.475\linewidth}
    \centering
    \begin{tabular}{lcc}
    \toprule
    \textbf{Method} & \textbf{1-shot} & \textbf{5-shot} \\
    \midrule
    Classifier-Baseline & 60.58 & 79.24 \\
    Classifier-Baseline (Euc.) & 56.29 & 78.93 \\
    \midrule
    Meta-Baseline & \textbf{63.33} & \textbf{80.02} \\
    Meta-Baseline (Euc.) & 60.19 & 79.50 \\
    \bottomrule
    \end{tabular}
    \caption{\textbf{Importance of inheriting a good metric.} Average 5-way accuracy (\%), with ResNet-12 on miniImageNet.}
    \label{tab:metric}
\end{minipage}
\vspace{-1em}
\end{table*}

\subsection{Results}
\label{std-benchmark}

Following the standard-setting, we conduct experiments on miniImageNet and tieredImageNet, the results are shown in Table~\ref{tab:mini} and \ref{tab:tiered} respectively. To get a fair comparison to prior works, we perform model selection according to the validation set. On both datasets, we observe that the Meta-Baseline achieves competitive performance to state-of-the-art methods. We highlight that \textit{many methods for comparison introduce more parameters and architecture designs (e.g. self-attention in \cite{xu2021constellation}), while Meta-Baseline has the minimum parameters and the simplest design.} We also notice that the simple Classifier-Baseline can achieve competitive performance when compared to meta-learning methods, especially in 5-shot tasks. We observe that the meta-learning stage consistently improves Classifier-Baseline on miniImageNet. Compared to miniImageNet, we find that the gap between Meta-Baseline and Classifier-Baseline is smaller on tieredImageNet, and the meta-learning stage does not improve 5-shot in this case.

We further evaluate our methods on the larger dataset ImageNet-800. In this larger-scale experiment, we find freezing the Batch Normalization layer~\cite{ioffe2015batch} (set to eval mode) is beneficial. The results are shown in Table~\ref{tab:im800}. From the results, we observe that in this large dataset Meta-Baseline improves Classifier-Baseline in 1-shot, while it is not improving the performance in 5-shot.

\section{Observations and Hypothesis}

\subsection{Objective discrepancy in meta-learning}

Despite the improvements of meta-learning over Classifier-Baseline, we observe the test performance drops during the meta-learning stage. While a common assumption for this phenomenon is overfitting, we observe that this issue seems not to be mitigated on larger datasets. To further locate the issue, we propose to evaluate \textit{base class generalization} and \textit{novel class generalization}. Base class generalization is measured by sampling tasks from unseen images in base classes, while novel class generalization refers to the performance of few-shot tasks sampled from novel classes. The base class generalization is the generalization in the input distribution for which the model is trained, it decouples the commonly defined generalization and class-level transfer performance, which helps for locating the reason for the performance drop.

Figure~\ref{fig:mbtr} and \ref{fig:mbtr-im800-mini} demonstrate the meta-learning stage of Meta-Baseline on different datasets. We find that during the meta-learning stage, when the base class generalization is increasing, the novel class generalization can be decreasing instead. This fact indicates that over a converged whole-classification model, the meta-learning objective itself, i.e. making the embedding generalize better in few-shot tasks from base classes, can have a negative effect on the performance of few-shot tasks from novel classes. It also gives a possible explanation for why such phenomenon is not mitigated on larger datasets, as this is not sample-level overfitting, but \textit{class-level overfitting}, which is caused by the objective discrepancy that the underlying training class distribution is different from testing class distribution.

This observation suggests that we may reconsider the motivation of the meta-learning framework for few-shot learning. In some settings, \textit{optimizing towards the training objective with a consistent form as the testing objective (except the inevitable class difference) may have an even negative effect}. It is also likely that the whole-classification learns the embedding with stronger class transferability, and meta-learning makes the model perform better at $N$-way $K$-shot tasks but tends to lose the class transferability.

\subsection{Effect of whole-classification training before meta-learning}
According to our hypothesis, the whole-classification pre-trained model has provided extra class transferability for the meta-learning model, therefore, it is natural to compare Meta-Baseline with and without the classification training stage. The results are shown in Table~\ref{tab:mbtrs}. We observe that Meta-Baseline trained without classification training stage can actually achieve higher base class generalization, but its novel class generalization is much lower when compared to Meta-Baseline with whole-classification training.

These results support our hypothesis, that the whole-classification training provides the embedding with stronger class transferability, which significantly helps novel class generalization. Interestingly, TADAM~\cite{oreshkin2018tadam} finds that co-training the meta-learning objective with a whole-classification task is beneficial, which may be potentially related to our hypothesis. While our results show it is likely that the key effect of the whole-classification objective is improving the class transferability, it also indicates a potential \textit{trade-off} that the whole-classification objective can have a negative effect on base class generalization.

\subsection{What makes Meta-Baseline a strong baseline?}

As a method with a similar objective as ProtoNet~\cite{snell2017prototypical}, Meta-Baseline achieves nearly 10\% higher accuracy on 1-shot in Table~\ref{tab:mini}. The observations and hypothesis in previous sections potentially explain its strength, as it starts with the embedding of a whole-classification model which has stronger class transferability.

We perform further experiments, that in Meta-Baseline (with classification training stage) we replace the cosine distance with the squared Euclidean distance proposed in ProtoNet~\cite{snell2017prototypical}. To get a fair comparison, we also include the learnable scalar $\tau$ with a proper initialization value $0.1$. The results are shown in Table~\ref{tab:metric}. While ProtoNet~\cite{snell2017prototypical} finds that squared Euclidean distance (as a Bregman divergence) works better than cosine distance when performing meta-learning from scratch, here we start meta-learning from Classifier-Baseline and we observe that cosine similarity works much better. A potential reason is that, as shown in Table~\ref{tab:metric}, cosine nearest-centroid works much better than nearest-centroid with squared Euclidean distance in Classifier-Baseline (note that this is just the evaluation metric and has no changes in training). \textit{Inheriting a good metric for Classifier-Baseline might be the key that makes Meta-Baseline strong.} According to our hypothesis, the embedding from the whole-classification model has strong class transferability, inheriting a good metric potentially minimizes the future modifications on the embedding from the whole-classification model, thus it can keep the class transferability better and achieve higher performance.

\subsection{Effect of dataset properties}

We construct four variants from the tieredImageNet dataset. Specifically, \textit{full-tiered} refers to the original tieredImageNet, \textit{full-shuffled} is constructed by randomly shuffling the classes in tieredImageNet and re-splitting the classes into training, validation, and test set. The \textit{mini-tiered} and \textit{mini-shuffled} datasets are constructed from full-tiered and full-shuffled respectively, their training set is constructed by randomly selecting 64 classes with 600 images from each class in the full training set, while the validation set and the test set remain unchanged. Since tieredImageNet separates training classes and testing classes into different super categories, shuffling these classes will mix the classes in different super categories together and make the distribution of base classes and novel classes closer.

Our previous experiments show that base class generalization is always improving, if novel classes are covered by the distribution of base classes, the novel class generalization should also keep increasing. From Table~\ref{tab:tvars}, we can see that from mini-tiered to mini-shuffled, and from full-tiered to full-shuffled, the improvement achieved by the meta-learning stage gets significantly larger, which consistently supports our hypothesis. Therefore, our results indicate it is likely that \textit{meta-learning is mostly effective over whole-classification training when novel classes are similar to base classes.}

We also observe that other factors may affect the improvement of meta-learning. From mini-tiered to full-tiered and from mini-shuffled to full-shuffled, when the dataset gets larger the improvements become less. A potential hypothesis could be that the class transferability advantage of whole-classification training becomes more obvious when trained on large datasets. From the results of our experiments in Table~\ref{tab:mini}, \ref{tab:tiered}, \ref{tab:im800}, \ref{tab:tvars}, we observe that the improvement of the meta-learning stage in 5-shot is less than 1-shot. We hypothesize this is because when there are more shots, taking average embedding becomes a more reasonable choice to estimate the class center in Classifier-Baseline, therefore the advantage of meta-learning becomes less.

\subsection{The trade-off between meta-learning and whole-classification}

All of the experiments in previous sections support a key hypothesis, that there exists a trade-off: the meta-learning objective learns better embedding for $N$-way $K$-shot tasks (in the same distribution), while the whole-classification objective learns embedding with stronger class transferability. Optimizing towards one objective may hurt the strength of another objective. With this hypothesis, Meta-Baseline balances this trade-off by choosing to calibrate the whole-classification embedding with meta-learning and inherit the metric with high initial performance.

The discrepancy between base class generalization and novel class generalization also considers the effectiveness of meta-learning and whole-classification from the perspective of datasets. Specifically, when novel classes are similar enough to base classes or the base classes are sufficient to cover the distribution of novel classes, novel class generalization should converge to base class generalization. In practice, this can be potentially achieved by collecting base classes that are similar to the target novel classes. In this case, it may be possible that the novel meta-learning algorithms outperform the whole-classification baselines again.

\begin{table*}
    \begin{center}
    \begin{tabular}{l|cc|ccc}
    \toprule
    \multicolumn{1}{c}{ } & \multicolumn{2}{c}{\textbf{Trained on ILSVRC}} & \multicolumn{3}{c}{\textbf{Trained on all datasets}} \\
    \multirow{2}{*}{Dataset} & \multirow{1}{*}{fo-Proto-MAML} & Classifier/Meta & \multirow{1}{*}{fo-Proto-MAML} & Classifier & Meta \\
    & \cite{triantafillou2019meta} & (ours) & \cite{triantafillou2019meta} & (ours) & (ours) \\
    \midrule
    ILSVRC & 49.5 & \textbf{59.2} & 46.5 & \textbf{55.0} & 48.0\\
    Omniglot & 63.4 & \textbf{69.1} & 82.7 & 76.9 & \textbf{89.4}\\
    Aircraft & \textbf{56.0} & 54.1 & 75.2 & 69.8 & \textbf{81.7}\\
    Birds & 68.7 & \textbf{77.3} & 69.9 & \textbf{78.3} & 77.3\\
    Textures & 66.5 & \textbf{76.0} & 68.3 & \textbf{71.4} & 64.5\\
    Quick Draw & 51.5 & \textbf{57.3} & 66.8 & 62.7 & \textbf{74.5}\\
    Fungi & 40.0 & \textbf{45.4} & 42.0 & 55.4 & \textbf{60.2}\\
    VGG Flower & 87.2 & \textbf{89.6} & 88.7 & \textbf{90.6} & 83.8\\
    Traffic Signs & 48.8 & \textbf{66.2} & 52.4 & \textbf{69.3} & 59.5\\
    MSCOCO & 43.7 & \textbf{55.7} & 41.7 & \textbf{53.1} & 43.6\\
    \bottomrule
    \end{tabular}
    \end{center}
    \caption{Additional results on Meta-Dataset. Average accuracy (\%), with variable number of ways and shots. The fo-Proto-MAML method is from Meta-Dataset~\cite{triantafillou2019meta}, Classifier and Meta refers to Classifier-Baseline and Meta-Baseline respectively, 1000 tasks are sampled for evaluating Classifier or Meta. Note that Traffic Signs and MSCOCO have no training set.}
    \label{tab:meta-dataset}
\end{table*}

\section{Additional Results on Meta-Dataset}

Meta-Dataset~\cite{triantafillou2019meta} is a new benchmark proposed for few-shot learning, it consists of diverse datasets for training and evaluation. They also propose to generate few-shot tasks with a variable number of ways and shots, for having a setting closer to the real world. We follow the setting in Meta-Dataset~\cite{triantafillou2019meta} and use ResNet-18 as the backbone, with the original image size of 126$\times$126, which is resized to be 128$\times$128 before feeding into the network. For the classification training stage, we apply the training setting similar to our setting in ImageNet-800. For the meta-learning stage, the model is trained for 5000 iterations with one task in each iteration.

The left side of Table~\ref{tab:meta-dataset} demonstrates the models trained with samples in ILSVRC-2012 only. We observe that the Meta-Baseline does not significantly improve Classifier-Baseline under this setting in our experiments, possibly due to the average number of shots are high.

The right side of Table~\ref{tab:meta-dataset} shows the results when the models are trained on all datasets, except Traffic Signs and MSCOCO which have no training samples. The Classifier-Baseline is trained as a multi-dataset classifier, i.e. an encoder together with multiple FC layers over the encoded feature to output the logits for different datasets. The classification training stage has the same number of iterations as training on ILSVRC only, to mimic the ILSVRC training, a batch has 0.5 probability to be from ILSVRC and 0.5 probability to be uniformly sampled from one of the other datasets. For Classifier-Baseline, comparing to the results on the left side of Table~\ref{tab:meta-dataset}, we observe that while the performance on ILSVRC is worse, the performances on other datasets are mostly improved due to having their samples in training. It can be also noticed that the cases where Meta-Baseline improves Classifier-Baseline are mostly on the datasets which are ``less relevant'' to ILSVRC (the dataset ``relevance'' could be shown in Dvornik et al.~\cite{dvornik2020selecting}). A potential reason is that the multi-dataset classification training stage samples ILSVRC with 0.5 probability, similar to ILSVRC training, the meta-learning stage is hard to improve on ILSVRC, therefore those datasets relevant to ILSVRC will have similar properties so that it is hard to improve on them.

\section{Conclusion and Discussion}

In this work, we presented a simple Meta-Baseline that has been overlooked for few-shot learning. Without any additional parameters or complex design choices, it is competitive to state-of-the-art methods on standard benchmarks.

Our experiments indicate that there might be an objective discrepancy in the meta-learning framework for few-shot learning, i.e. a meta-learning model generalizing better on unseen tasks from base classes might have worse performance on tasks from novel classes. This provides a possible explanation that why some complex meta-learning methods could not get significantly better performance than simple whole-classification. While most recent works focus on improving the meta-learning structures, many of them did not explicitly address the issue of class transferability. Our observations suggest that the objective discrepancy might be a potential key challenge to tackle.

While many novel meta-learning algorithms are proposed and some recent works report that simple whole-classification training is good enough for few-shot learning, we show that meta-learning is still effective over whole-classification models. We observe a potential trade-off between the objectives of meta-learning and whole-classification. From the perspective of datasets, we demonstrate how the preference between meta-learning and whole-classification changes according to class similarity and other factors, indicating that these factors may need more attention for model comparisons in future work.

~

{\footnotesize \textbf{Acknowledgements.}~This work was supported, in part, by grants from DARPA LwLL, NSF 1730158 CI-New: Cognitive Hardware and Software Ecosystem Community Infrastructure (CHASE-CI), NSF ACI-1541349 CC*DNI Pacific Research Platform, and gifts from Qualcomm, TuSimple and Picsart. Prof. Darrell was supported, in part, by DoD including DARPA's XAI, LwLL, and/or SemaFor programs, as well as BAIR's industrial alliance programs. We thank Hang Gao for the helpful discussions.}


{\small
\bibliographystyle{ieee_fullname}
\bibliography{main}
}

\clearpage

\appendix

\begin{figure}
    \begin{center}
        \includegraphics[width=\linewidth]{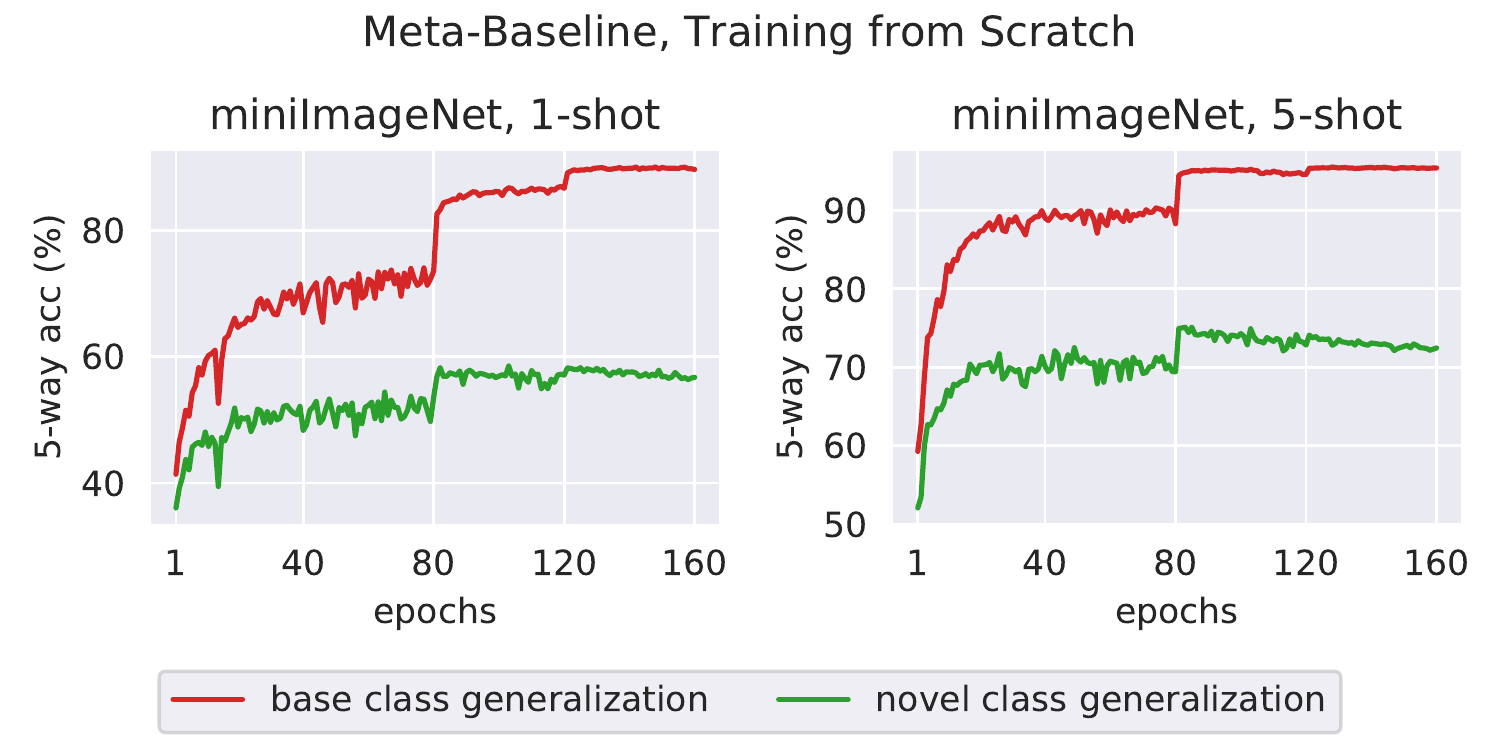}
    \end{center}
    \caption{Training Meta-Baseline without classification-training stage on miniImageNet.}
    \label{fig:mbtrs}
\end{figure}

\begin{table}
    \begin{center}
        \begin{tabular}{llcc}
            \toprule
            \textbf{Dataset} & \textbf{Classifier} & \textbf{1-shot} & \textbf{5-shot} \\
            \midrule
            \multirow{2}{*}{miniImageNet} & Linear & 60.58 & \textbf{79.24} \\
            & Cosine & \textbf{61.93} & 78.73 \\
            \midrule
            \multirow{2}{*}{tieredImageNet} & Linear & \textbf{68.76} & \textbf{84.07} \\
            & Cosine & 67.58 & 83.31 \\
            \bottomrule
        \end{tabular}
    \end{center}
    \caption{Comparison to classifier trained with cosine metric, Average 5-way accuracy (\%), with ResNet-12.}
    \label{tab:pretrain-cos}
\end{table}

\begin{figure*}
    \begin{center}
        \includegraphics[width=\linewidth]{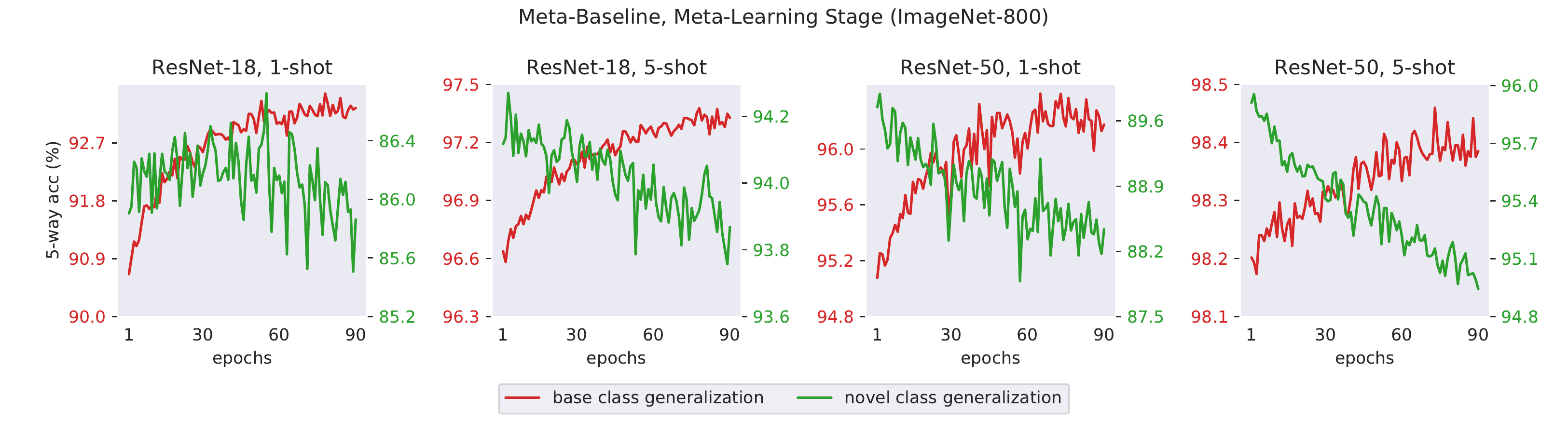}
    \end{center}
    \vspace{-1em}
    \caption{\textbf{Objective discrepancy of meta-learning on ImageNet-800.} Each epoch contains 500 training batches. Average 5-way accuracy (\%) is reported.}
    \label{fig:mbtr-im800}
\end{figure*}

\begin{table*}
    \begin{center}
        \begin{tabular}{cc}
            \toprule
            \textbf{difference} & \textbf{5-way 1-shot accuracy (\%)} \\
            \midrule
            (default, reported~\cite{chen2018a}) & 51.75 $\pm$ 0.80 \\
            (default, reproduced by their code) & 50.84 $\pm$ 0.80 \\
            finetune $\rightarrow$ cosine nearest-centroid & 52.15 $\pm$ 0.83 \\
            epoch-300 $\rightarrow$ epoch-50 & 53.37 $\pm$ 0.71 \\
            remove color-jittering & 56.06 $\pm$ 0.71 \\
            224$\times$224 input size $\rightarrow$ resizing 84$\times$84 to 224$\times$224 & 50.49 $\pm$ 0.71 \\
            ResNet-18 $\rightarrow$ ResNet-12 & 53.59 $\pm$ 0.72 \\
            Adam (lr=0.001, batch=16) $\rightarrow$ SGD (lr=0.1, batch=128) & 59.19 $\pm$ 0.71 \\
            \bottomrule
        \end{tabular}
    \end{center}
    \vspace{-1em}
    \caption{Comparison of the Classifier-Baseline and Baseline++~\cite{chen2018a}.}
    \label{tab:abl-closer}
\end{table*}

\section{Details of ResNet-12}

The ResNet-12 backbone consists of 4 residual blocks that each residual block has 3 convolutional layers. Each convolutional layer has a $3\times 3$ kernel, followed by Batch Normalization~\cite{ioffe2015batch} and Leaky ReLU~\cite{xu2015empirical} with 0.1 slope. The channels of convolutional layers in each residual block are 64, 128, 256, 512 respectively, a $2\times 2$ max-pooling layer is applied after each residual block. Finally, a $5 \times 5$ global average pooling is applied to get a 512-dimensional feature vector.

This architecture is consistent with recent works~\cite{oreshkin2018tadam,xu2021constellation}. Some other recent works also introduce additional parameters and design choices in the backbone (e.g. DropBlock and wider channels of 64, 160, 320, 640 in \cite{lee2019meta,tian2020rethink}), while these modifications may make the performance higher, we do not include them here for simplicity.

\section{Training plot of Meta-Baseline without classification training stage}

We show the process of training Meta-Baseline from scratch (i.e. without the classification-training stage) on miniImageNet in Figure~\ref{fig:mbtrs}. We observe that when the learning rate decays, the novel class generalization quickly starts to be decreasing. While it is able to achieve higher base class generalization than Meta-Baseline with classification training, its highest novel class generalization is still much worse, suggesting whole-classification training may provide representations with extra class transferability.

\section{Comparison to cosine classification training}

We compare the effect of classification training with replacing the last linear-classifier with cosine nearest-neighbor metric which is proposed in prior work~\cite{gidaris2018dynamic,chen2018a}, the results are shown in Table~\ref{tab:pretrain-cos}, where Cosine denotes classification training with cosine metric and Linear denotes the standard classification training. On miniImageNet, we observe that Cosine outperforms Linear in 1-shot, but has worse performance in 5-shot. On tieredImageNet, we observe Linear outperforms Cosine in both 1-shot and 5-shot. We choose to use the linear layer as it is more common and we find it works better in more cases.

\section{Objective discrepancy on ImageNet-800}

Besides miniImageNet and tieredImageNet, in our large-scale dataset ImageNet-800, we also observe the novel class generalization decreasing when base class generalization is increasing, the training process is demonstrated in Figure~\ref{fig:mbtr-im800}. From the figure, we see that for both backbones of ResNet-18 and ResNet-50, the base class generalization performance is increasing during the training epochs, while the novel class generalization performance quickly starts to be decreasing. These observations are consistent with our observations on miniImageNet and tieredImageNet, which further support our hypothesis.

\section{Comparison of the Classifier-Baseline and Baseline++~\cite{chen2018a}}

We connect the Classifier-Baseline to Baseline++~\cite{chen2018a} with a step-by-step ablation study on miniImageNet, the results are shown in Table~\ref{tab:abl-closer}. We see that fine-tuning is outperformed by the simple nearest-centroid method with cosine metric, and using a standard ImageNet-like optimizer significantly improves the performance of the whole-classification method for few-shot learning.

\end{document}